\documentclass[runningheads]{llncs}
\usepackage{graphicx}
\usepackage{subcaption}

\usepackage{pgfplots}
\usepackage[ruled,vlined]{algorithm2e}
\usepackage{amsmath,amssymb,amsfonts}
\usepackage{algorithmic}
\usepackage{makecell,multirow}
\usepackage{enumitem}
\setlist{nolistsep}

\makeatletter
\g@addto@macro\normalsize{%
\setlength\abovedisplayskip{-6pt}
\setlength\belowdisplayskip{0pt}
\setlength\abovedisplayshortskip{-6pt}
\setlength\belowdisplayshortskip{0pt}
}
\makeatother
\begin{document}
\title{Kernel-Level Energy-Efficient Neural Architecture Search for Tabular Dataset \thanks{ This work was supported in part by European Commission under MISO project (grant 101086541), EEA grant under the HAPADS project (grant NOR/POLNOR/HAPADS/0049/2019-00), Research Council of Norway under eX3 project (grant 270053) and Sigma2 (grant NN9342K).}}
%
%
\author{Hoang-Loc La\thanks{Corresponding author}\orcidID{0009-0005-5453-7836} \and Phuong Hoai Ha\orcidID{0000-0001-8366-5590}}
\authorrunning{H-L La et al.}

\institute{The Arctic University of Norway, Norway \\
\email{\{hoang.l.la,phuong.hoai.ha\}@uit.no} }

\maketitle 
\vspace*{-\baselineskip}
\begin{abstract}
Many studies estimate energy consumption using proxy metrics like memory usage, FLOPs, and inference latency, with the assumption that reducing these metrics will also lower energy consumption in neural networks. This paper, however, takes a different approach by introducing an energy-efficient Neural Architecture Search (NAS) method that directly focuses on identifying architectures that minimize energy consumption while maintaining acceptable accuracy. Unlike previous methods that primarily target vision and language tasks, the approach proposed here specifically addresses tabular datasets. Remarkably, the optimal architecture suggested by this method can reduce energy consumption by up to 92\% compared to architectures recommended by conventional NAS.

\keywords{Neural Architecture Search \and Energy-Efficient NAS\and Energy Consumption Prediction.}
\vspace*{-\baselineskip}
\end{abstract}

\section{Introduction}
\label{Sect:introduction}
\vspace{-0.5cm}
\par Tabular datasets are among the oldest and most widely used types of datasets in practice, appearing in various fields such as medicine, finance, environmental science, and more. Alongside tree-based machine learning techniques, neural networks are a popular method for tackling tasks involving tabular data. However, as neural network models grow more complex, they demand more hardware resources, leading to higher energy consumption. To address this challenge, energy-efficient deep learning has emerged as a viable solution.
\par Very large neural models that achieve state-of-the-art accuracy are heavily dependent on the computational and memory capabilities of hardware, which become a big problem on mobile devices and edge platforms. To address this challenge, several approaches have been introduced to discover optimal neural architectures using hardware-aware metrics, including ProxylessNAS \cite{cai2018proxylessnas} and MnasNet \cite{tan2019mnasnet}. Nonetheless, these methods mainly consider proxy metrics, such as memory usage and latency, while assuming a strong correlation between the energy consumption of neural networks and these metrics.
\par In a different approach, several studies have focused on directly minimizing the energy consumption of neural networks by utilizing neural architecture search (NAS) techniques. Bakhtiarifard et al. \cite{bakhtiarifard2024ec} published a benchmark dataset that captures the energy usage of various architectures during the inference phase on NVIDIA GPUs. Their goal is to search for optimal architectures from a set of architectures, considering both accuracy and energy efficiency. They employed multiple NAS techniques to discover these optimal solutions. However, their methodology requires profiling all candidate models on the target hardware, which becomes impractical when dealing with a large number of candidate architectures. Additionally, if new search spaces that are not included in the benchmark dataset are introduced, it would be necessary to re-profile and assess the energy consumption of models derived from them.
\par Unlike previous model-level approaches, our proposed energy-efficient NAS employs a kernel-level energy consumption predictor, which can be easily adapted to any neural architecture. Specifically, our energy prediction model is inspired by nn-meter \cite{zhang2021nn}, a well-known kernel-level latency predictor for neural networks. However, unlike latency profiling, measuring energy consumption is more challenging. We provide a detailed explanation of how we profile energy consumption on NVIDIA Jetson boards in Section \ref{sub:energy_profiling}. Moreover, the original algorithm in \cite{zhang2021nn} does not account for the parallelism present in NVIDIA GPUs. To address this, we propose an enhanced algorithm that fills this gap, making the nn-meter compatible with both desktop and edge NVIDIA GPUs. In summary, our main contributions are as follows.
\begin{itemize} 
    \item We propose an enhanced method for accurately predicting the energy consumption and inference latency of neural networks on NVIDIA GPUs. Our approach complements the method introduced in nn-meter \cite{zhang2021nn}. 
    \item Unlike previous works that primarily focus on vision tasks, we introduce an energy-efficient NAS method specifically designed for tabular tasks. Vision tasks typically take 2-D data as input, whereas tabular tasks use 1-D vectors, necessitating a different architectural approach. We employ three tailored search spaces: MLP, ResNet, and FTTransformer \cite{gorishniy2021revisiting}. It is remarkable that we use MLP-style ResNet, which replace convolution layers of the original ResNet architecture with Fully Connected layers. To the best of our knowledge, this is the first work to propose an energy-efficient NAS for tabular tasks.
    \item We conduct extensive experiments to emphasize the importance of integrating energy consumption considerations into the NAS process, which can help lower the energy usage of neural networks in deployment environments.
\end{itemize}

\vspace{-0.3cm}

\section{Related Work}
\label{Sect:related_work}
\vspace{-0.3cm}
\subsection{Energy Prediction for Neural Networks}
\vspace{-0.2cm}
\par Estimating energy consumption of neural networks during inference has been addressed in several studies. Yang et al. \cite{yang2017method} developed an energy model primarily based on Floating Point Operations (FLOPs) and Memory-Access Counts (MACs) at both cache and DRAM levels. They predicted the FLOPs and MACs of individual neural layers on specific hardware using simulation. However, with the release of new hardware platforms and optimization techniques, building such a simulator to accurately estimate FLOPs and MACs has become impractical.
\par In a different approach, the authors of \cite{cai2017neuralpower} introduced NeuralPower, a layer-wise energy consumption predictor for neural networks. This method assumes that the energy consumption of each layer is independent and that all layers execute sequentially. The total energy consumption of the model is then calculated by summing the energy usage of individual layers. However, due to recent software optimization techniques, such as layer fusion \cite{tensorrt_documentation} and computation graph optimization, neural network layers can now be fused or executed in parallel, invalidating NeuralPower's assumptions. A similar assumption was also made in \cite{lahmer2022energy}, which proposes an analytic energy consumption model for Convolutional Networks on NVIDIA's Jetson board \cite{lahmer2022energy}.
\par To overcome the above problem, nn-Meter \cite{zhang2021nn} introduced a kernel-level latency predictor designed for various hardware platforms and deep learning libraries. Their approach begins by conducting experiments on the target backend to identify fusion kernels. They then create a dataset containing the latency measurements of these kernels on the backend. Based on this dataset, they build latency predictors for each kernel. When predicting the latency of a neural network, the network is first split into fused kernels using a kernel detection algorithm. The latency of each kernel is predicted using the pre-trained predictors, and the total latency of the neural network is calculated as the sum of the latencies of all the kernels. This concept was further adapted to energy consumption by Tu et al. \cite{tu2023unveiling} for mobile platforms, making it the most relevant work to our study.
\par Unlike previous research, which primarily focuses on convolutional networks for mobile devices, this paper extends the approach to the NVIDIA Jetson device family, specifically targeting tabular networks. Notably, measuring GPU power consumption on NVIDIA Jetson boards poses significant challenges \cite{holly2020profiling}, which we will elaborate on in Section \ref{sub:energy_profiling}. Additionally, the kernel detection algorithm used in these earlier works assumes that all kernels run sequentially. However, as we demonstrate in Section \ref{sub:parallelable}, this assumption is incorrect, and we propose an improved algorithm to address this limitation in the kernel detection process.
\vspace{-0.2cm}
\subsection{Neural Architecture Search}
\vspace{-0.2cm}
\subsubsection{One-shot Neural Architecture Search}
\par In conventional NAS, to evaluate candidate models' performance, each candidate architecture is typically trained until it converges, a process that can be extremely time-consuming when the search space is large. To address this, several approaches have been developed to bypass the training phase using performance estimators. One such approach is one-shot NAS. One-shot NAS builds a supernet that encompasses the entire search space, where every edge in the supernet represents all possible operations that can be assigned to it. Notably, architectures that share a specific operation also share the corresponding weights, enabling simultaneous training of a vast number of subnetworks. As a result, one-shot NAS can reduce GPU training time by up to 1000x compared to the tranditional NAS. Candidate architectures are sampled from the supernet, and their accuracy can either be evaluated directly or with minimal fine-tuning (few-shot NAS).
\par The concept of weight-sharing was first introduced in ENAS \cite{pham2018efficient}, where a supernet was proposed to cover multiple candidate architectures. Instead of training each architecture individually, ENAS shares parameters across these architectures, significantly reducing the total training time and resource consumption. Notable works in this direction include DARTS \cite{liu2018darts}, SPOS \cite{guo2020single}. These methods generally share weights among subnets during supernet training while decoupling the weights of different operations within the same layer. However, applying these weight-sharing techniques directly to transformer-based search spaces presents training challenges for the supernet.

\par To address this issue, Autoformer \cite{sukthanker2023weight} introduced a technique called weight-entanglement, which reduces memory consumption and improves the convergence of supernet training. The key idea is to allow different transformer blocks to share weights for common components within each layer. The weight-entanglement strategy ensures that different candidate blocks in the same layer share as many weights as possible. A similar approach has also been explored for convolutional networks \cite{wang2021mergenas}, \cite{sukthanker2023weight}.

\par Unlike previous works, which focus mainly on vision tasks, this paper targets tabular tasks and adapts the weight-entanglement concept to three different search spaces: MLP-based space, ResNet-based space, and FTTransformer-based space. 

\vspace{-0.5cm}
\subsubsection{Energy-Efficient Neural Architecture Search}

\par In addition to accuracy and latency, energy consumption has become a critical factor when selecting optimal neural architectures for deployment environments. Early work on energy-efficient NAS primarily relied on approximate computing techniques, such as reducing memory usage \cite{han2015learning} or leveraging layer sparsity \cite{he2018amc}, to lower energy consumption.

\par On the other hand, several studies have directly incorporated energy consumption into the NAS process.  ETNAS \cite{dong2023etnas} uses accuracy as a constraint while focusing on minimizing the average power consumption of architectures on NVIDIA Desktop GPUs. In contrast, Bakhtiarifard et al. \cite{bakhtiarifard2024ec} treated energy-efficient NAS as a multi-objective optimization (MOO) problem, applying several MOO evolutionary algorithms to find a Pareto-front of optimal architectures. Similarly, Sukthanker et al. \cite{sukthanker2024multi} introduced MODNAS, a one-shot NAS framework designed for various target devices with multiple hardware metrics. To predict energy consumption, MODNAS employed the HELP technique \cite{lee2021hardware} to meta-learn energy predictors across multiple devices. This approach is particularly useful when an energy consumption dataset exists for several target devices, allowing for quick adaptation of energy predictors to new devices with minimal additional training data. One problem of MODNAS method is that they requires profiling energy usage of a subset of the considered search space on a set of multiple devices to transfer-learn the energy predictors. Their method is beneficial when quickly adapting for a new device with an acceptable accuracy. However, when adapting for a new search space, the method needs to collect new measurements with the new search space. On another hand, our method can accurately predict energy consumption for arbitrary search spaces without any further data collection process.

\par Similar to the ETNAS approach, our NAS algorithm also treats energy-efficient NAS as an accuracy-constrained problem. However, unlike ETNAS, we directly optimize for accurate energy consumption rather than average power consumption. Additionally, while search space of ETNAS was developed for vision tasks, our proposed search spaces are specifically designed for tabular tasks.

\vspace{-0.3cm}
\section{Kernel-based Energy Model}
\label{Sect:power_predictor}

\vspace{-0.3cm}
\subsection{Overall ideas}
\vspace{-0.2cm}
\par Our energy model is inspired by the approach used in nn-Meter \cite{zhang2021nn}. The latency predictor in \cite{zhang2021nn} is based on the observation that fused kernels in a neural network are executed sequentially. Specifically, nn-Meter first applies a heuristic method to detect fusion rules for a given hardware platform and deep learning backend. It then represents a neural architecture as a graph and uses a breadth-first search algorithm to traverse all nodes, merging multiple operator nodes into fused kernels according to the detected rules. The overall latency is then calculated as the sum of the latencies of these fused kernels. To predict kernel-level latency, nn-Meter builds latency predictors that take kernel parameters as inputs. One limitation of nn-Meter's kernel splitting strategy is that it does not account for parallelism of kernel execution on NVIDIA GPUs, which we discuss further in Section \ref{sub:parallelable}.

\par We adapt these ideas for our kernel-level energy model. However, unlike latency, which can be measured accurately on current devices with nanosecond precision, energy profiling is constrained by the sampling frequency of power measurement tools. For example, a typical convolution layer on the Jetson Orin completes in a few hundred microseconds, while the default configuration of the onboard power sensor takes around 1.4 milliseconds per sample. Direct energy profiling under these conditions results in unstable measurements and, consequently, reduced accuracy of the energy model.

\par
\begin{figure}[!htb]
    \centering
    \includegraphics[width=0.87\textwidth]{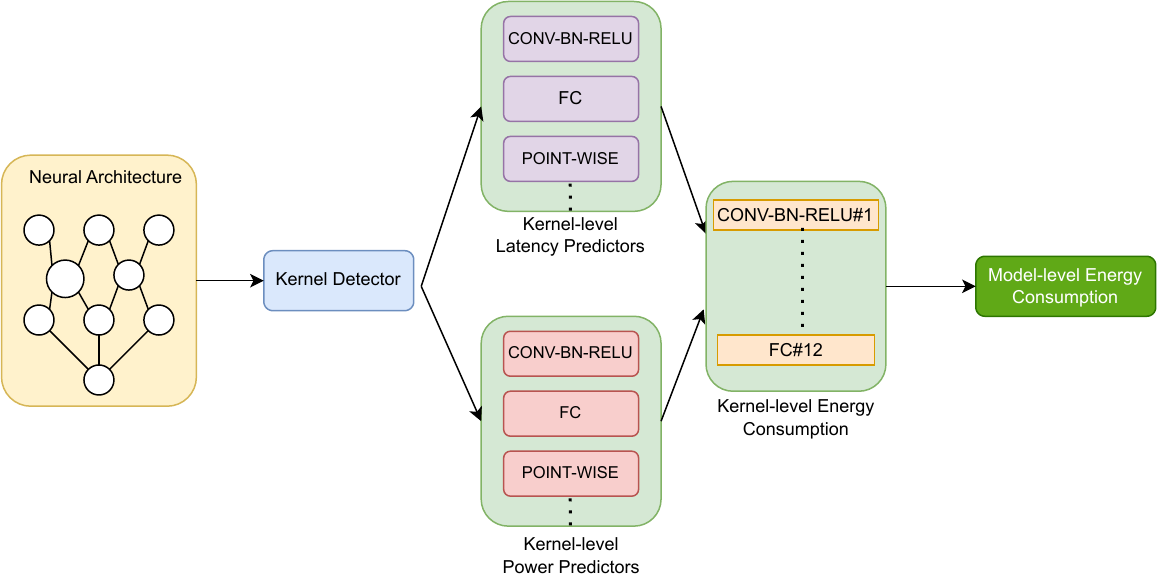}
    \vspace{-0.3cm}
    \caption{Kernel-based Energy Model.}
    \label{fig:energy_model}
    \vspace{0.2cm}
\end{figure}

 On the other hand, profiling the average power consumption of kernels is more feasible and stable. Instead of developing kernel-level predictors for energy consumption directly, we use power consumption as the basis for our model. Our kernel-level energy model, illustrated in Figure \ref{fig:energy_model}, predicts both power consumption and latency for each kernel. The energy consumption at the kernel level is simply the product of the predicted latency and power consumption, and the total energy consumption for the model is the sum of the energy consumption across all kernels.

\vspace{-0.3cm}
\subsection{Parallelizable Kernels on NVIDIA GPUs}
\vspace{-0.3cm}
\label{sub:parallelable}
\par Beside the fusion mechanism, NVIDIA GPUs are able to run several convolution-based kernels with the same configurations and input matrices in parallel. In Section \ref{sub:micro-benchmark}, we conducted experiments with micro-benchmarks to verify this observation. Notably, the maximum number of kernels that can be run in parallel varies across different NVIDIA GPUs. For example, the NVIDIA Jetson AGX can support up to 8 parallel kernels, while the NVIDIA Quadro RTX4000 can handle up to 16. 
\par
\begin{figure}[!htb]
    
    \vspace{-0.8cm}
        \centering
        \includegraphics[width=\textwidth]{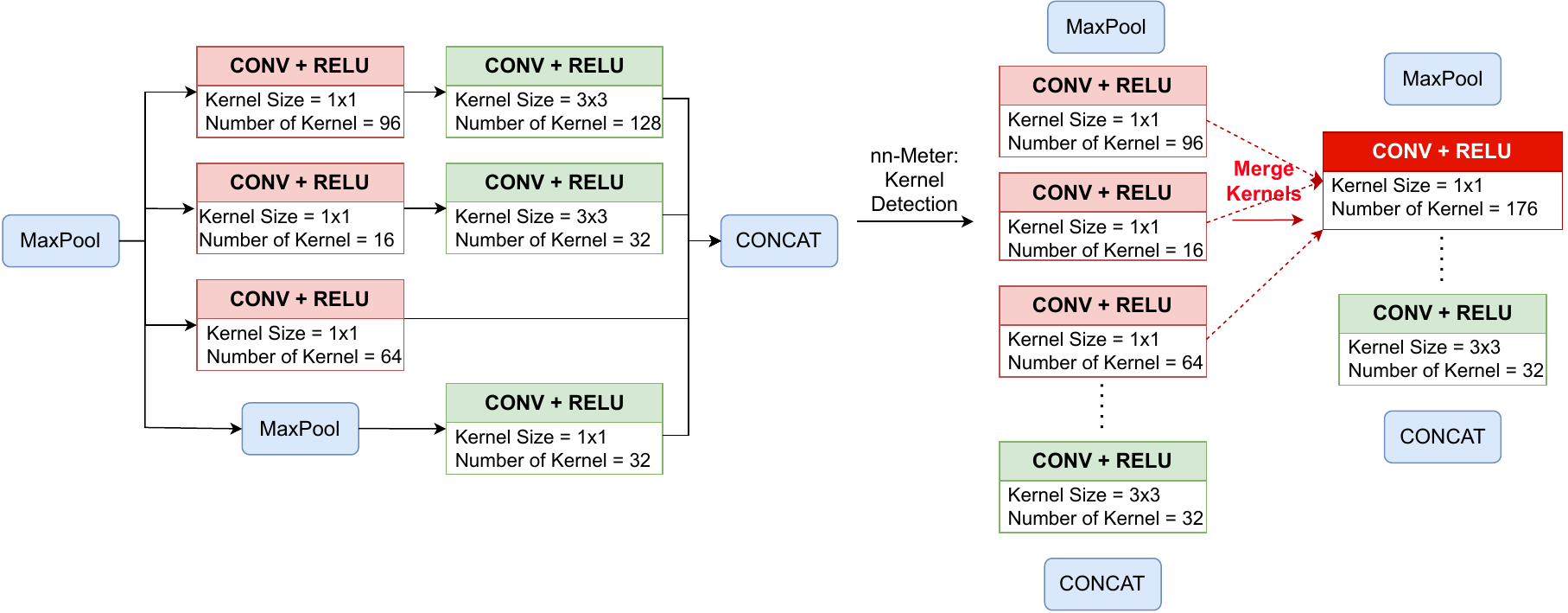}
        \vspace{-0.4cm}
        \caption{Examples of parallelizable kernels from GoogLeNet\cite{szegedy2015going}. The light red rectangles denote parallelizable kernels. The red rectangle denotes the newly generated kernels}
        \label{fig:parallelable_example}
\vspace{-0.2cm}
\end{figure}

 The kernel detection algorithms used in \cite{zhang2021nn} and \cite{tu2023unveiling} do not take this parallelism into account. Figure \ref{fig:parallelable_example} illustrates an example of parallelizable convolution kernels from GoogLeNet\cite{szegedy2015going}. This architectural pattern is found across the Inception family \cite{szegedy2015going,szegedy2016rethinking,szegedy2017inception} for vision tasks. Incorporating this parallelism mechanism can lead to more accurate performance and energy consumption predictions on NVIDIA GPUs.

\vspace{-0.2cm}

\par To extend the kernel detection algorithm \cite{zhang2021nn} for parallizeable kernels, we propose Algoritm \ref{alg:cap1}. Our proposed algorithm applies a Breadth-First Search (BFS) traversal to identify all kernels at the same level within the fused computation graph, which is derived from the original kernel detection algorithm. It's important to note that fused kernels often consist of multiple conventional layers. For example, a fused kernel like conv+bn+relu includes three layers: convolution, batch normalization, and ReLU. 
\par After identifying these fused kernels, we group convolution kernels that share the same kernel size, strides, number of groups, and dilation rate. These grouped kernels will now have the same kernel size, although they may differ in the number of filters. For each group, we generate a new convolution kernels with the same configuration as the original convolutions, but the number of filters will be the sum of the filters from all kernels in the group. 
\par To predict the energy consumption and latency of these parallelizable kernels, we use this newly generated kernels. Figure \ref{fig:parallelable_example} illustrates the process of creating these new convolution kernels. This method allows for accurate prediction of performance metrics by considering the parallelism capabilities of the hardware.
\vspace{-1.1cm}
\begin{algorithm}[!htb]\scriptsize
\caption{Kernel Splitting for Parallelizable Kernels}\label{alg:cap1}
\begin{algorithmic}[1]
\REQUIRE $G$ is a fused graph, which is a results of Kernel Detection algorithm \cite{zhang2021nn}.
\STATE Denote a queue $Q$ containing all node at the current depth
\STATE Denote a dictionary $in\_degree$ contains number of incoming edges for each node.
\STATE Initialize  $Q$ and $in\_degree$.
    \WHILE{$Q$ is not empty}
        \STATE $current\_level$ is a list of all node in the current level. \\
        \textit{\# BFS traversal to group all kernels, which are at the same level}\\
        \FOR{$N_{cur} \in Q$}
            \STATE $N_{cur} = Q.dequeue()$
            \STATE $current\_level.append(N_{cur})$
            \FOR {$N_{succ} \in N_{cur}.out$}
                \STATE $in\_degree[N_{succ}] -= 1$
                \IF {$in\_degree[N_{succ}] == 0$}
                    \STATE $Q.enqueue(N_{succ})$
                \ENDIF
            \ENDFOR
        \ENDFOR
        \STATE Remove non-convolutional kernels from $current\_level$ list.
        \STATE Group kernels from the list based on their type and configurations.
        \STATE Merge kernels at the same group by generating a new kernel.
    \ENDWHILE
\end{algorithmic}
\end{algorithm}

\vspace{-1.4cm}
\subsection{Power Profiling Method for NVIDIA Jetson boards}
\label{sub:energy_profiling}
\vspace{-.1cm}
\par NVIDIA Jetson boards monitor power consumption using three-channel INA3221 sensors. When measuring power consumption on these boards, Burtscher et al. \cite{burtscher2014measuring} observed unexpected behaviors from the built-in sensors and hypothesized that capacitor charging and discharging on the board caused these anomalies. However, this behavior is actually due to the accumulation register within the sensor itself \cite{ina3221_documentation}. The capacitor charging effect they observed is essentially the Moving Average Value computed by the accumulation register. This was empirically confirmed by applying a Moving Average Filter to the raw data from the sensor, as demonstrated in the study by Aslan et al. \cite{aslan2022study}.
\par Remarkably, the sampling speed of INA3221 sensors depend on two factors, namely the clock frequency of $i^{2}c$ protocol and register configuration of INA3221. To overcome the inaccurate problem of built-in INA3221 sensors, we adjust the register configuration of reading the INA3221 sensors by reducing the conversion time to minimum \cite{ina3221_documentation} and re-compile OS kernel of Jetson board to increase the clock frequency of $i^{2}c$ protocol from the default value (400KHz) to the maximum value (1Mhz).

\par
\begin{figure}[!htb]
    \centering
    \includegraphics[width=0.85\textwidth]{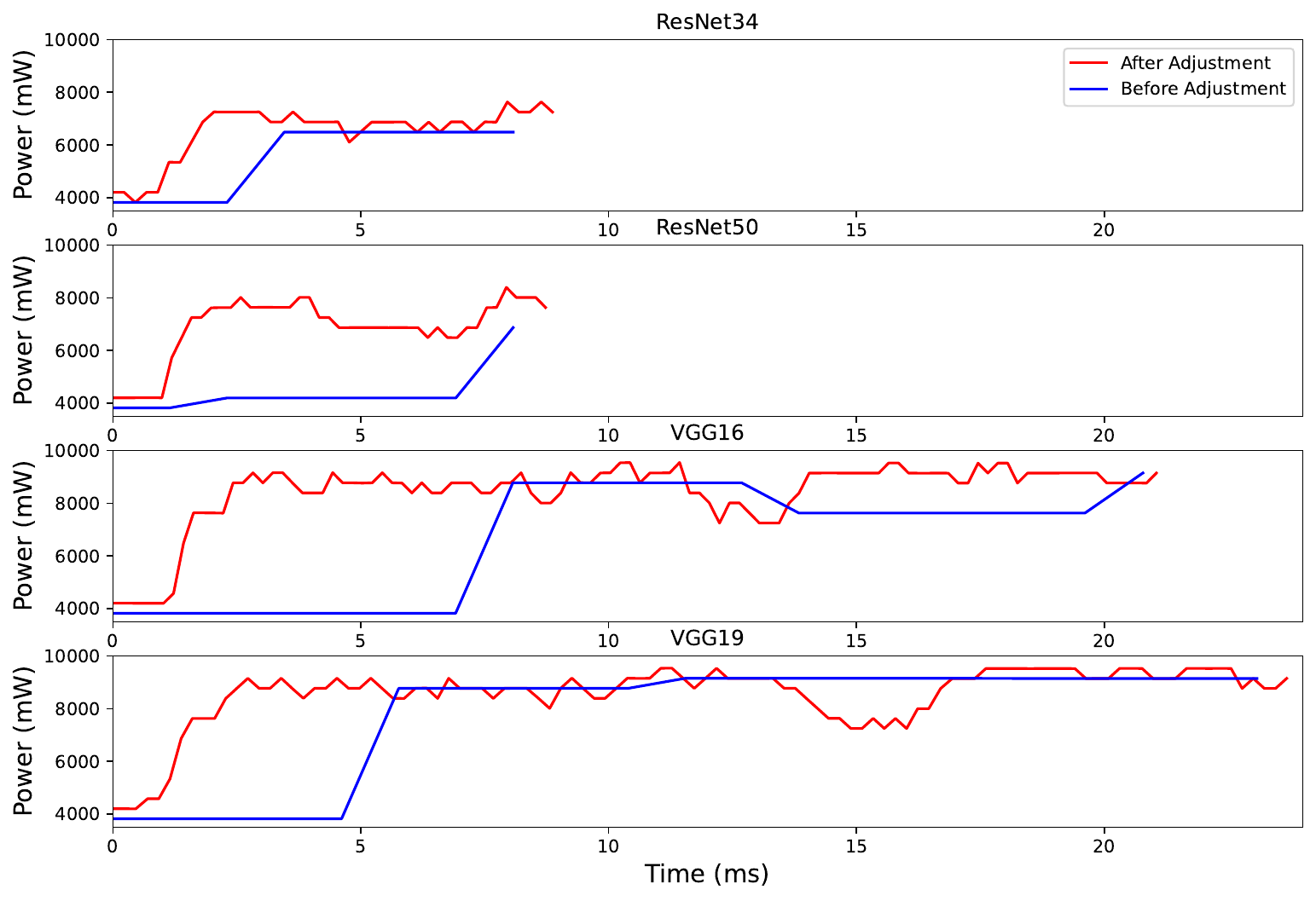}
    \vspace{-0.3cm}
      \caption{Measured power consumption of common CNNs on the Jetson AGX Orin before and after adjustments, with lines ending when inference stops.}
      \label{fig:sampling_method}
\vspace{-1cm}
\end{figure}

 Figure \ref{fig:sampling_method} illustrates the differences in power readings from the built-in sensors before and after the adjustments. The data indicates a significant discrepancy between the two measurements. Before the adjustments, the power readings were inaccurate; for instance, although the network began at the first time step, the blue line only started to increase after several milliseconds. Additionally, during inference, the GPU executed various neural kernels, which should have resulted in changes to power consumption, yet the blue line remained almost flat.
\par In contrast, after adjusting the INA3221 configuration and increasing the $i^{2}c$ frequency, the sensors accurately tracked power consumption trends during neural network execution. The red line shows an immediate increase when the network runs, reflecting fluctuations in power usage throughout the execution.


\vspace{-0.8cm}

\section{Energy-Efficient Neural Architecture Search}
\label{Sect:nas}

\vspace{-0.3cm}
\subsection{Define search space}

\vspace{-0.2cm}
\par Motivated by previous research on deep learning for tabular datasets \cite{mcelfresh2024neural,huang2020tabtransformer,gorishniy2021revisiting}, we propose three search spaces based on distinct backbones: Multi-Layer Perceptron (MLP), ResNet, and FTTransformer \cite{gorishniy2021revisiting}. The possible configurations for each search space are detailed in Table \ref{tab:2}. Tuples of three values in parentheses represent the lowest value, highest, and steps. 
\vspace{-1cm}
\begin{table}[!htb]\scriptsize
    \centering
    \caption{POSSIBLE CONFIGURATIONS OF THREE SEARCH SPACES.}
    \begin{tabular}{|c|c|c|c|}
        \hline
         & \textbf{FTTransformer} & \textbf{ResNet} & \textbf{MLP}  \\ 
        \hline
        \multirow{5}{*}{\makecell[c]{\textbf{Possible} \\ \textbf{Choices}}} & \# of Blocks: (1, 8, 1) & \# of Blocks: (1, 11, 1) & \# of Blocks: (1, 11, 1) \\
         & \# of Heads: (2, 8, 1) & Hidden Dim: (16, 512, 16) & Hidden Dim: (16, 512, 16) \\
         & Embedded Dim: (16, 256, 16) & Backbone Dim: (16, 512, 16) & \\ 
         & Q-K-V Dim: (16, 256, 16) & &  \\
         & MLP Ratio: (1.0, 4.0, 0.5) & &  \\
        \hline
         \makecell[c]{\textbf{\# of} \\ \textbf{Candidates}}& $(7*16*16*7)^{8}$ & $32 * 32^{11}$& $32^{11}$ \\
         \hline
    \end{tabular}
    \label{tab:2}
\end{table}

\vspace{-0.5cm}
\par
\begin{figure}[!htb]
    \centering 

    \begin{subfigure}[b]{0.46\textwidth} 
        \includegraphics[width=\textwidth]{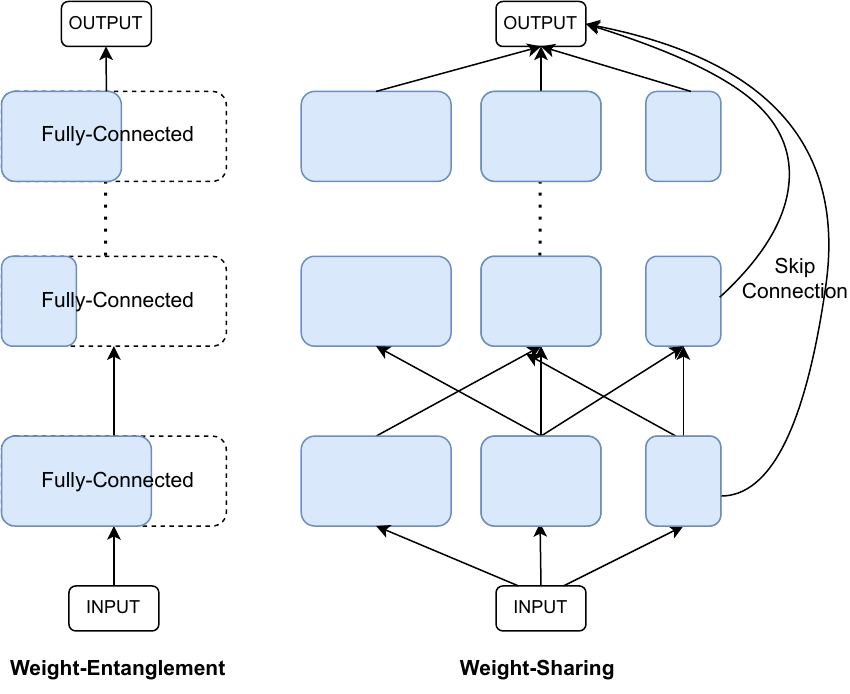}
        \caption{\textit{Left}}
        \label{fig:supernet_2}
    \end{subfigure}
    \hfill 
    \begin{subfigure}[b]{0.46\textwidth} 
        \includegraphics[width=\textwidth]{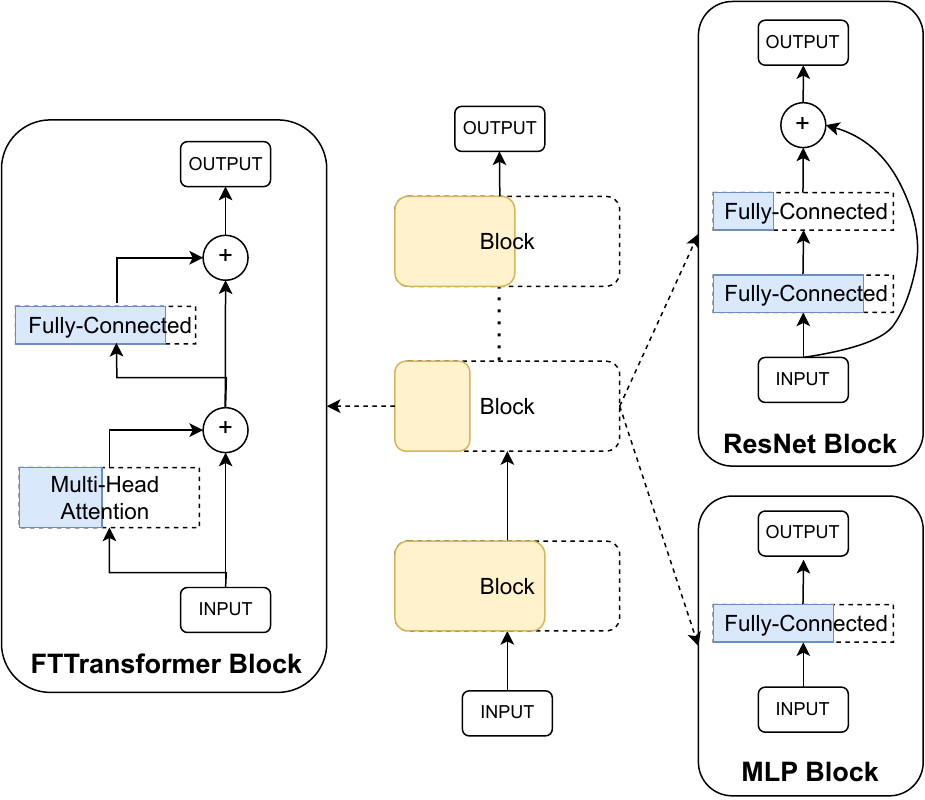}
        \caption{\textit{Right}}
        \label{fig:supernet_1}
    \end{subfigure}
    \caption{\textit{Left}: The difference between the weight-entanglement supernet and weight-sharing supernet for MLP search space. \textit{Right}: Overall architecture of the three supernets for tabular search spaces, showing selected components in solid lines and unselected components in dashed lines. All supernets share the same macro backbone, with differences in block configurations.}
    \label{fig:supernet_comparison}
    
\vspace{-0.9cm}
\end{figure}

 Traditional Neural Architecture Search (NAS) methods require training each candidate model from scratch, which is time-consuming. To address this, we adapt a one-shot method by constructing a supernet for each search space. Building on the weight-entanglement concept from Autoformer \cite{chen2021autoformer}, we introduce three weight-entanglement supernets. Figure \ref{fig:supernet_2} illustrates the key difference between the weight-entanglement supernet and the weight-sharing one for the MLP search space. It is similar for the ResNet and FTTransformer search spaces. In the weight-sharing schema, the weights of all candidate blocks at the same layer are decoupled. In contrast, the weight-entanglement schema allows these weights to be shared among all candidate blocks.
\par Figure \ref{fig:supernet_1} presents the overall architecture of the three tabular supernets. Notably, the configuration of each block and the number of blocks within these architectures are dynamic.

\vspace{-0.9cm}

\vspace{-0.1cm}
\subsection{Searching Strategy}
\vspace{-0.2cm}
\par In this paper, we focus on minimizing the energy consumption of neural networks while maintaining a soft constraint on accuracy. To facilitate the NAS process in finding optimal solutions, we employ a Policy-Gradient-based Reinforcement Learning algorithm \cite{sutton1999policy}. Let $s$ as a candidate architecture generated from a supernet $S$. The term $Energy(s)$ and $Accuracy(s)$ denote energy consumption and accuracy of the subnet $s$, respectively, while $T$ represents the target accuracy. The optimization problem is defined as in Equation (\ref{eq:1}).

\begin{equation}
\begin{aligned}
\label{eq:1}
&\underset{s \in S }{\text{minimize}}
& &Energy(s) * [\frac{Accuracy(s)}{T}]^{w} \\
\end{aligned}
\end{equation}

\par With the trade-off factor $w$ defined as:\\
\begin{equation}
\begin{aligned}
&  w = \begin{cases} 
\alpha, & \text{if } Accuracy(m) \leq T \\
\beta, & \text{otherwise}
\end{cases}
\end{aligned}
\end{equation}
\par The trade-off between accuracy and energy can be adjusted by changing $\alpha$ and $\beta$. Empirically, we choose $\alpha=-2$ and $\beta=-1/2$, which implies that when $Accuracy$ is below the threshold $T$, the reward function becomes exponentially inversely proportional to the accuracy. If the accuracy exceeds this threshold, the reward function is less sensitive to accuracy and places more emphasis on the energy term.

\vspace{-0.5cm}
\section{Experiments}
\label{Sect:Experiments}
\vspace{-0.3cm}
\subsection{Experimental Setup}
\vspace{-0.2cm}
\par We utilize nine regression datasets from the TabZilla benchmark \cite{mcelfresh2024neural} along with one practical dataset from a real-world application. TabZilla is a prominent benchmark for tabular data, offering a diverse range of datasets for both regression and classification tasks. Our emphasis is primarily on regression, as the real-world dataset also pertains to a regression problem. The considered datasets from Tabzilla are Bank-Note-Authentication-UCI (BNA-UCI), california, cpu-small, dataset\_sales, EgyptianSkulls, kin8nm, liver-disorders, mv, and Wine. For real-world application, we leverage MISO dataset, which consists of air quality measurements collected from a network of low-cost sensors, in conjunction with ground-truth data obtained from a reliable reference station. The main objective is to calibrate the readings from the low-cost sensors to align with the ground-truth values. Specifically, for each dataset, we split it into training and testing parts with fractions of 40\% and 60\%, respectively. 
\par 


\par Additionally, our energy model supports three platforms: the NVIDIA Jetson Nano, NVIDIA Jetson AGX Orin, and the Intel Neural Network Stick 2 (NCS2) \cite{intel_ncs2}. To measure the energy consumption on the NVIDIA Jetson boards, we utilize built-in sensors alongside the methods outlined in Section \ref{sub:energy_profiling}. The deep learning framework used is NVIDIA TensorRT. For the Intel NCS2, we employ an external Monsoon Power Monitor \cite{monsoon_power_monitor} to profile energy consumption. 


\vspace{-0.2cm}
\subsection{Energy Prediction Model}
\subsubsection{Parallelizable Kernels}
\label{sub:micro-benchmark}
\par
\begin{figure}[!htb]
\centering
		\includegraphics[width=0.4\linewidth]{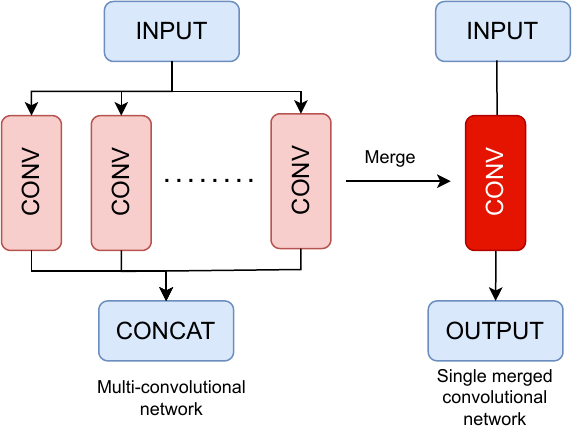}
		\label{fig:parallel_latency_1}
        
        \vspace{-0.2cm}
     \caption{Overview of the multi-convolution network and the single merged convolution network used in our micro-benchmark.}
    \label{fig:micro_benchmark}
\end{figure}

 We conduct empirical experiments with micro-benchmarks to validate the proposed algorithm outlined in Section \ref{sub:parallelable}. First, we construct a neural network with multiple convolution layers that take the same input, with their outputs merged by a concatenation layer. Next, we create a single merged convolution network by merging all parallelizable convolution layers. Finally, we compare the inference latency and energy consumption of the generated single merged convolution network with that of the multi-convolution network, assessing both parallel and sequential execution of     the convolution layers. Figure \ref{fig:micro_benchmark} depicts the network topology of these convolution networks. For the latency metric, we perform experiments on two different NVIDIA GPU platforms: the NVIDIA Jetson AGX Orin and the NVIDIA Quadro RTX4000. Energy consumption measurements are conducted exclusively on the NVIDIA Jetson AGX Orin. 

\par
\begin{figure}[!htb]
\centering
    \begin{subfigure}{0.7\linewidth}
		\includegraphics[width=\linewidth]{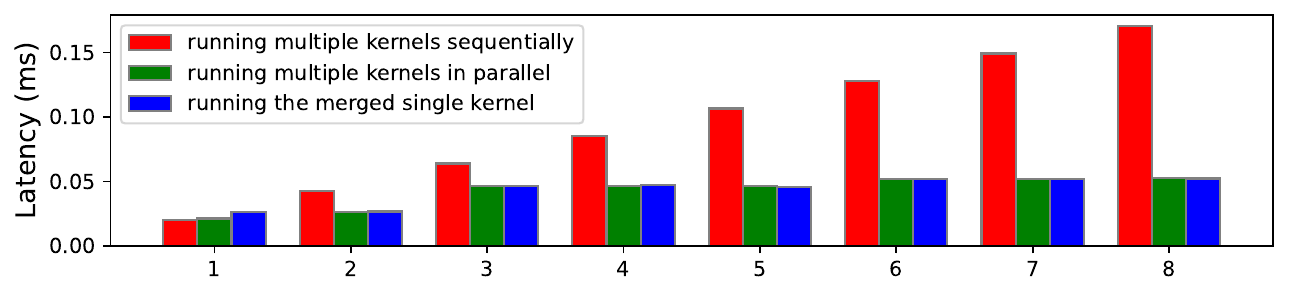}
		\label{fig:parallel_latency_1}
	   \end{subfigure}
    \begin{subfigure}{0.7\linewidth}
		\includegraphics[width=\linewidth]{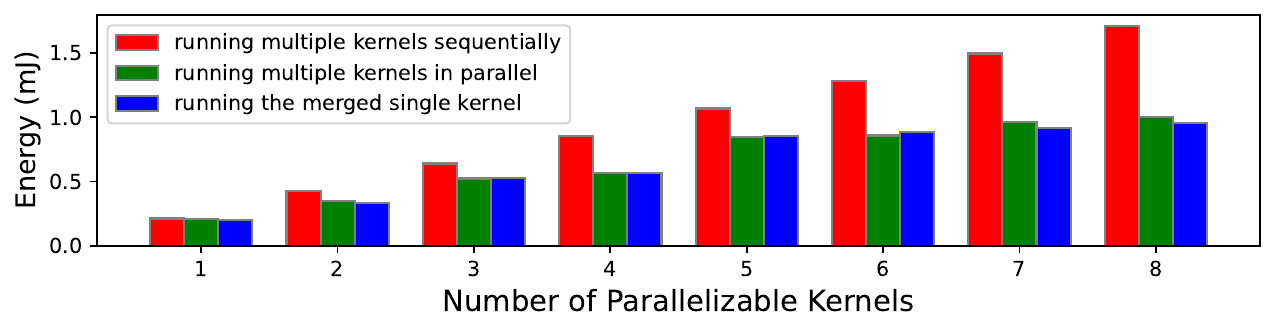}
		\caption{NVIDIA Jetson AGX Orin}
        \vspace{0.2cm}
		\label{fig:parallelabl_energy}
	   \end{subfigure}
    \vfill
    \begin{subfigure}{0.7\linewidth}
		\includegraphics[width=\linewidth]{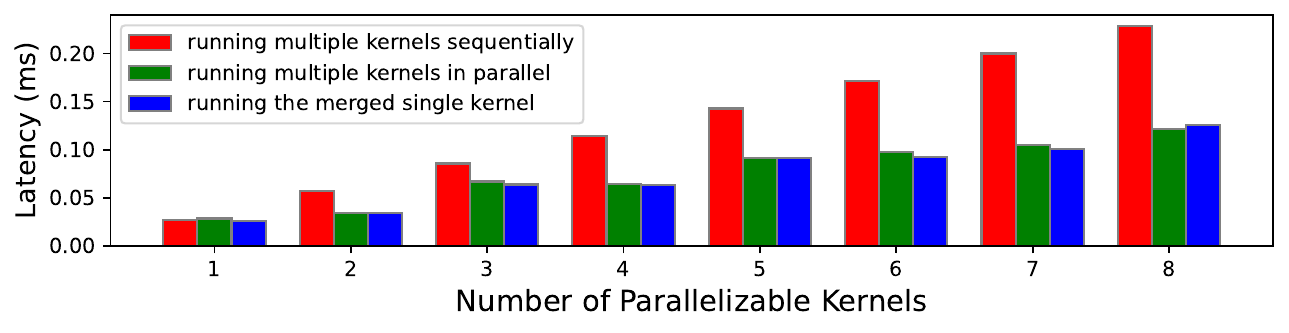}
        
        \vspace{-0.2cm}
		\caption{NVIDIA Quadro RTX4000}
		\label{fig:parallel_latency_2}
        
        \vspace{-0.2cm}
	   \end{subfigure}
     \caption{Inference latency and energy consumption (measured only on AGX Orin) across two NVIDIA GPU platforms for the multi-convolution network with kernels executed sequentially or in parallel, compared to the single-convolution network generated as outlined in Section \ref{sub:parallelable}.}
    \label{fig:parallel_latency}
    
\end{figure}

 Figure \ref{fig:parallel_latency} presents the experimental results. We observe significant differences in inference time and energy consumption between executing the convolution layers in parallel versus sequentially. Additionally, the latency and energy consumption of the merged convolution layer are comparable to those of the convolution layers executed in parallel.
\subsubsection{End-to-end Prediction}
\par We assess the accuracy of our energy prediction model in an end-to-end setting. First, we create a benchmark of the 10 most popular CNNs: convolution-style ResNet, AlexNet, DenseNet, GoogLeNet, InceptionV3, SqueezeNet, Inception+ResNet, MnasNet, ShuffleNet, and MobileNetV2. Next, we randomly generate 20 different candidate models from each search space. Finally, we compare the actual energy consumption of these models on the Jetson Nano and Intel NCS2 with the predicted values of our energy model. Table \ref{tab:3} shows the results of the experiments. Our energy prediction model demonstrates high accuracy for  CNNs, MLPs,  our MLP-style ResNets, and FTTransformer. For FTTransformer-based models, TensorRT utilizes Myelin library to compile and optimize graph computations. Myelin supports intensive pointwise fusions, which is particularly advantageous for transformer-like models. Although our current energy model also supports such fusions, the specific implementation details of Myelin remain a black box. 
      
\vspace{-1cm}
\begin{table}[!htb]\scriptsize
    \centering
    \caption{ACCURACY OF END-TO-END PREDICTIONS ON JETSON NANO AND INTEL NCS2}
    \begin{tabular}{|c|c|c|c|c|}
        \hline
        \multirow{ 2}{*}{\textbf{Benchmark}}   & \multicolumn{2}{c|}{\textbf{Jetson Nano}} & \multicolumn{2}{c|}{\textbf{Intel NCS2}}   \\
        \cline{2-5}
        & Latency & Energy & Latency & Energy   \\ 
        \hline
        CNNs & 0.965 & 0.9002 & 0.912 & 0.91\\ 
        \hline
        MLP & 0.978 & 0.925 & 0.956 & 0.934\\ 
        \hline
        ResNet & 0.968 & 0.935 & 0.964 & 0.912  \\
        \hline
        FTTransformer& 0.894 & 0.871 & Unsupported &  Unsupported\\
        \hline
    \end{tabular}
    \label{tab:3}
    
\end{table}      
\vspace{-1cm}

\subsection{Energy-efficient NAS}
\par To evaluate the energy efficiency of architectures identified by our energy-efficient NAS, we compare them with those found through conventional NAS (which focuses solely on accuracy) in terms of both accuracy and energy consumption on the NVIDIA Jetson Nano. Specifically, we employ the $R^{2}$ Score to assess the accuracy of regression tasks.
\par We also compare our energy-efficient NAS with another similar method, ETNAS, which is also an energy-aware NAS. They also use policy-based reinforcement method for the neural searching stage. However, unlike our approach, ETNAS aims to minimize total power consumption across all layers of the network. Their original method, though, is designed for a vision-specific search space, which isn't applicable to tabular tasks. To ensure a fair comparison, we adapt ETNAS method with our tabular search spaces, denoted as Adapted-ETNAS. Table \ref{tab:4} shows a comparison between our proposed NAS with two other baselines, namely Conventional NAS, and Adapted ETNAS. To facilitate clearer comparison, we use the energy-saving metric, which indicates the reduction in energy consumption achieved by applying the model suggested by our proposed NAS relative to the model produced by the conventional approach. Particualarly, we train all supernets with the training set and evaluate the accuracy of optimal architectures proposed by above methods with the testing part.

\par Importantly, both energy-efficient NAS methods identify architectures that substantially enhance energy efficiency while maintaining accuracy levels comparable to those recommended by conventional NAS. Additionally, we find that no single search space consistently yields the highest accuracy across all datasets, as model performance varies based on dataset characteristics. However, our proposed NAS consistently discovers the most energy-efficient architectures compared to ETNAS, with optimal architectures achieving up to 91.9\% energy savings over conventional NAS. Although the difference in energy consumption between our approach and ETNAS is relatively minor in the MLP search space, it becomes more significant in the FTTransformer and ResNet search spaces, particularly for the MISO and liver-disorders datasets. The small gaps in the MLP search space can be attributed to its simplicity, whereas the more complex FTTransformer and ResNet architectures reveal greater differences in energy efficiency. Furthermore, ETNAS considers only power consumption and disregards inference latency, a critical factor that affects the overall energy consumption of neural networks.

\begin{table*}[t]\scriptsize
    \centering
    \caption{A COMPARISION BETWEEN THE PROPOSED METHOD WITH OTHER BASELINES}
    \label{tab:4}
    \begin{tabular}{|c|c| c|c | c|c | c| c| c|}
    \hline
    \multirow{4}{*}{\textbf{Benchmark}} & \multirow{4}{*}{\textbf{Dataset}} & \multicolumn{7}{c|}{\textbf{FTTransformer}}  \\
    \cline{3-9}
     &  & \multicolumn{2}{c|}{ Conventional NAS} &  \multicolumn{2}{c|}{Adapted-ETNAS} & \multicolumn{3}{c|}{\textbf{Proposed NAS}} \\
    \cline{3-9}
    & & \makecell[c]{Energy \\ (mJ)}  & Accuracy & \makecell[c]{Energy \\ (mJ)}  & Accuracy & \makecell[c]{Energy \\ (mJ)} & Accuracy & \makecell[c]{Energy \\ Saving (\%)}\\
    \hline
    \multirow{9}{*}{\textbf{Tabzilla}} & \makecell[c]{BNA-UCI} & 13.214 & 1.0 & 1.393 & 1.0 & 1.350 & 1.0 & 89.8 \\ 
    \cline{2-9}
     & \makecell[c]{california} &  5.563 & 0.697 & 1.924 & 0.621 & 1.398 & 0.691 & 74.9\\
    \cline{2-9}
     & \makecell[c]{cpu\_small} & 5.561 & 0.967 & 2.001 & 0.963 & 1.550 & 0.95 & 72.1 \\
    \cline{2-9}
     & \makecell[c]{dataset\_sales} &  5.562 & 0.809 & 1.702 & 0.776 & 1.636 & 0.701 & 70.6\\
    \cline{2-9}
     & \makecell[c]{EgyptianSkulls} & 5.6 & 0.096 & 1.55 & 0.023 & 1.350 & 0.061 & 75.9\\
    \cline{2-9}
     & \makecell[c]{kin8nm} & 3.826 & 0.941 & 1.403 & 0.923 & 1.392 & 0.921 & 82.8 \\
    \cline{2-9}
     & \makecell[c]{liver-disorders} & 5.578 & 0.292 & 1.477 & 0.234 & 1.387 & 0.243 & 75.1 \\
    \cline{2-9}
     & \makecell[c]{mv} & 3.011 & 1.0 & 1.503 & 1.0 & 1.372 & 1.0 & 54.4 \\
    \cline{2-9}
     & \makecell[c]{Wine} & 5.569 & 0,338 & 1.744 & 0.173 & 1.550 & 0.216 & 72.2 \\
    \hline
     \multirow{1}{*}{\textbf{Real Use-Case}}& \makecell[c]{MISO} & 8.114 & 0.99 & 3.504 & 0.985 & 2.810 & 0.987 & 65.4 \\
    \hline
    \end{tabular}
    \begin{tabular}{|c|c| c|c | c|c | c| c| c|}
    \hline
    \multirow{4}{*}{\textbf{Benchmark}} & \multirow{4}{*}{\textbf{Dataset}} & \multicolumn{7}{c|}{\textbf{ResNet}}  \\
    \cline{3-9}
     &  & \multicolumn{2}{c|}{ Conventional NAS} &  \multicolumn{2}{c|}{Adapted-ETNAS} & \multicolumn{3}{c|}{\textbf{Proposed NAS}} \\
    \cline{3-9}
    & & \makecell[c]{Energy \\ (mJ)}  & Accuracy & \makecell[c]{Energy \\ (mJ)}  & Accuracy & \makecell[c]{Energy \\ (mJ)} & Accuracy & \makecell[c]{Energy \\ Saving (\%)}\\
    \hline 
    \multirow{9}{*}{\textbf{Tabzilla}} & \makecell[c]{BNA-UCI}  & 9.354 & 1.0 & 0.891 & 0.997 & 0.866 & 0.997 & 90.7\\ 
    \cline{2-9}
    & \makecell[c]{california}  & 6.08 & 0.799 & 0.981 & 0.674 & 0.863 & 0.782 & 85.8\\
    \cline{2-9}
    & \makecell[c]{cpu\_small} & 10.035 & 0.969 & 0.988 & 0.917 & 0.867 & 0.927 & 91.4\\
    \cline{2-9}
    & \makecell[c]{dataset\_sales} & 10.056 & 0.734 & 0.907 & 0.673 & 0.868 & 0.685 & 91.4\\
    \cline{2-9}
    & \makecell[c]{EgyptianSkulls} &  10.709 & 0,288 & 1.911 & 0.134 & 0.87 & 0.205 & 91.9\\
    \cline{2-9}
    & \makecell[c]{kin8nm} & 2.819 & 0.917 & 0.897 & 0.914 & 0.867 & 0.906 & 69.3\\
    \cline{2-9}
    & \makecell[c]{liver-disorders} & 9.339 & 0.315 & 1.838 & 0.239 & 0.881 & 0.228 & 90.6\\
    \cline{2-9}
    & \makecell[c]{mv} & 8.452 & 1.0 & 0.877 & 1.0 & 0.866 & 0.998 & 89.8\\
    \cline{2-9}
    & \makecell[c]{Wine} & 7.179 & 0.396 & 0.917 & 0.373 & 0.867 & 0.374 & 87.9\\
    \hline
    \multirow{1}{*}{\textbf{Real Use-Case}} & \makecell[c]{MISO} & 6.933 & 0.986 & 0.887 & 0.984 & 0.866 & 0.982 & 87.5\\
    \hline
    \end{tabular}
    \begin{tabular}{|c|c| c|c | c|c | c| c| c|}
    \hline
    \multirow{4}{*}{\textbf{Benchmark}} & \multirow{4}{*}{\textbf{Dataset}} & \multicolumn{7}{c|}{\textbf{MLP}}  \\
    \cline{3-9}
     &  & \multicolumn{2}{c|}{ Conventional NAS} &  \multicolumn{2}{c|}{Adapted-ETNAS} & \multicolumn{3}{c|}{\textbf{Proposed NAS}} \\
    \cline{3-9}
    & & \makecell[c]{Energy \\ (mJ)}  & Accuracy & \makecell[c]{Energy \\ (mJ)}  & Accuracy & \makecell[c]{Energy \\ (mJ)} & Accuracy & \makecell[c]{Energy \\ Saving (\%)}\\
    \hline 
    \multirow{9}{*}{\textbf{Tabzilla}} & \makecell[c]{BNA-UCI} & 5,191 & 1.0 & 0.533 & 0.905 & 0.520 & 0.998 & 90.0\\ 
    \cline{2-9}
    & \makecell[c]{california} & 5.092 & 0.675 & 0.533 & 0.624 & 0.521 & 0.668 & 89.8 \\
    \cline{2-9}
    & \makecell[c]{cpu\_small} & 5.139 & 0.922 & 0.528 & 0.769 & 0.519 & 0.75  & 89.9\\
    \cline{2-9}
    & \makecell[c]{dataset\_sales} & 3.313 & 0.65 & 0.532 & 0.677 & 0.521 & 0.658 & 84.3\\
    \cline{2-9}
    & \makecell[c]{EgyptianSkulls} & 3.288 & 0.288 &0.531  & 0.265 & 0.524 & 0.299 & 84.1\\
    \cline{2-9}
    & \makecell[c]{kin8nm} & 2.36 & 0.937 & 0.531 & 0.926 & 0.523 & 0.918 & 77.8\\
    \cline{2-9}
    & \makecell[c]{liver-disorders} & 3.3 & 0.221 & 0.531 & 0.209 & 0.524 & 0.219 & 84.1\\
    \cline{2-9}
    & \makecell[c]{mv} & 1.366 & 1.0 & 0.533 & 1.0 & 0.518 & 0.997 & 62.0\\
    \cline{2-9}
    & \makecell[c]{Wine} & 2.381 & 0.384 & 0.533 & 0.373 & 0.523 & 0.351 & 78.0\\
    \hline
    \multirow{1}{*}{\textbf{Real Use-Case}} & \makecell[c]{MISO} & 1.578 & 0.989 & 0.533 & 0.905 & 0.518 & 0.983 & 67.2\\
    \hline
    \end{tabular}
\end{table*}
      
\vspace{-0.2cm} 

\vspace{-0.2cm}
\section{Conclusions and Future Work}
\label{Sect:Conclusion}
\vspace{-0.3cm}
In this paper, we propose a energy-efficient NAS leveraging a kernel-level energy prediction model. Our energy-efficient NAS can be easily adapted for new search space without requiring further data collection. The proposed NAS can search optimal architectures in terms of energy efficiency with comparable accuracy. One current problem with our energy-efficient NAS is that when adapting for new devices, we need to re-collect/re-profile energy consumption on this new platform. In future, we will leverage meta-learning techniques to streamline this cumbersome data collection process, 

\bibliographystyle{splncs04}
\bibliography{Refs}


\end{document}